\documentclass[11pt]{article}
\usepackage{caption}
\usepackage{verbatimbox}
\usepackage[nocompress]{cite}
\usepackage{qtree}
\usepackage{color}
\usepackage{times}
\usepackage{epsfig}
\usepackage{graphicx}
\usepackage{amssymb}
\usepackage{amsmath}
\usepackage{amssymb}
\usepackage{url}
\usepackage{placeins}
\usepackage{caption}
\usepackage{multicol}
\usepackage[margin=1in]{geometry}
\usepackage{subfig}
\usepackage{stfloats}
\usepackage{placeins}
\usepackage{paralist, tabularx}

\usepackage{xcolor}

\title{SANTIAGO: Spine Association for Neuron Topology Improvement and Graph Optimization}

\author{William Gray Roncal$^{1,2}$, Colin Lea$^{1}$, 
Akira Baruah$^{3}$, Gregory D. Hager$^{1}$\\
$^{1}$Johns Hopkins University, Department of Computer Science, Baltimore, MD\\
$^{2}$JHU Applied Physics Laboratory, Research and Exploratory Development, Laurel, MD \\
$^{3}$ Columbia University, New York, NY \\
\noindent\rule[0.5ex]{0.3\linewidth}{1pt} \\
Corresponding Author:  William Gray Roncal\\ email: wgr@jhu.edu }

\newcommand{\websitesantiago}{\url{github.com/wrgr/santiago}}

\definecolor{willcolor}{RGB}{23, 127, 117}
\definecolor{bcolor}{RGB}{182, 33, 45}

\begin{document}

\maketitle

\begin{abstract} 

Developing automated and semi-automated solutions for reconstructing wiring diagrams of the brain from electron micrographs is important for advancing the field of connectomics.  While the ultimate goal is to generate a graph of neuron connectivity, most prior automated methods have focused on volume segmentation rather than explicit graph estimation. In these approaches, one of the key, commonly occurring error modes is dendritic shaft-spine fragmentation.

We posit that directly addressing this problem of \textit{connection identification} may provide critical insight into estimating more accurate brain graphs.  To this end, we develop a network-centric approach motivated by biological priors image grammars.  We build a computer vision pipeline to reconnect fragmented spines to their parent dendrites using both fully-automated and semi-automated approaches.  Our experiments show we can learn valid connections despite uncertain segmentation paths.  We curate the first known reference dataset for analyzing the performance of various spine-shaft algorithms and demonstrate promising results that recover many previously lost connections.  Our automated approach improves the local subgraph score by more than four times and the full graph score by 60 percent.  These data, results, and evaluation tools are all available to the broader scientific community. This reframing of the connectomics problem illustrates a semantic, biologically inspired solution to remedy a major problem with neuron tracking. 

\end{abstract}
\section{Introduction}

In adult vertebrate brains, each spine is typically associated with an excitatory synaptic connection, making their detection and association a critical task for building brain graphs \cite{Arellano2007,Yuste2010}.  Many current algorithms in connectomics are designed and evaluated using surrogate metrics (e.g. voxel-level segmentation of neurites)  rather than global graph measures.  Therefore, even the best available segmentation, coupled with ground truth synapses, may produce a poor estimate of connectivity.  Current results are quite accurate at capturing large process segmentation; one large contributor to network degradation in vertebrate brains is spine neck fragmentation. This is caused by  small cross-sectional areas, densely packed structures, limited contrast, and poor overlap due to anisotropy.  
Biologically, spines are small projections from the dendritic shafts of neuronal cells.  Spines occur predominantly in vertebrates and are prolific -- a single human brain likely contains many billions of these structures.  These structures are difficult to track in existing imaging methods due to their small length (a few microns) and volume (1 femtoliter) \cite{Yuste2010}.  The cross-sectional area of spine necks are typically $\approx 0.2$ microns \cite{Yuste2010}, corresponding to only a few pixels across a single imaging plane at the resolution typically used in serial section electron microscopy.   Spines were discovered by Santiago Ramon y Cajal \cite{RamonyCajal}, back in the late 19th century and it is hypothesized that understanding their function will unlock many of the secrets of neuronal computation \cite{Yuste2010}.  

\begin{figure}[htbp!!]
    \centering
    \begin{minipage}{.65\textwidth}
        \centering
        \hspace{-0.4in}\includegraphics[width=1\linewidth, height=0.45\textheight]{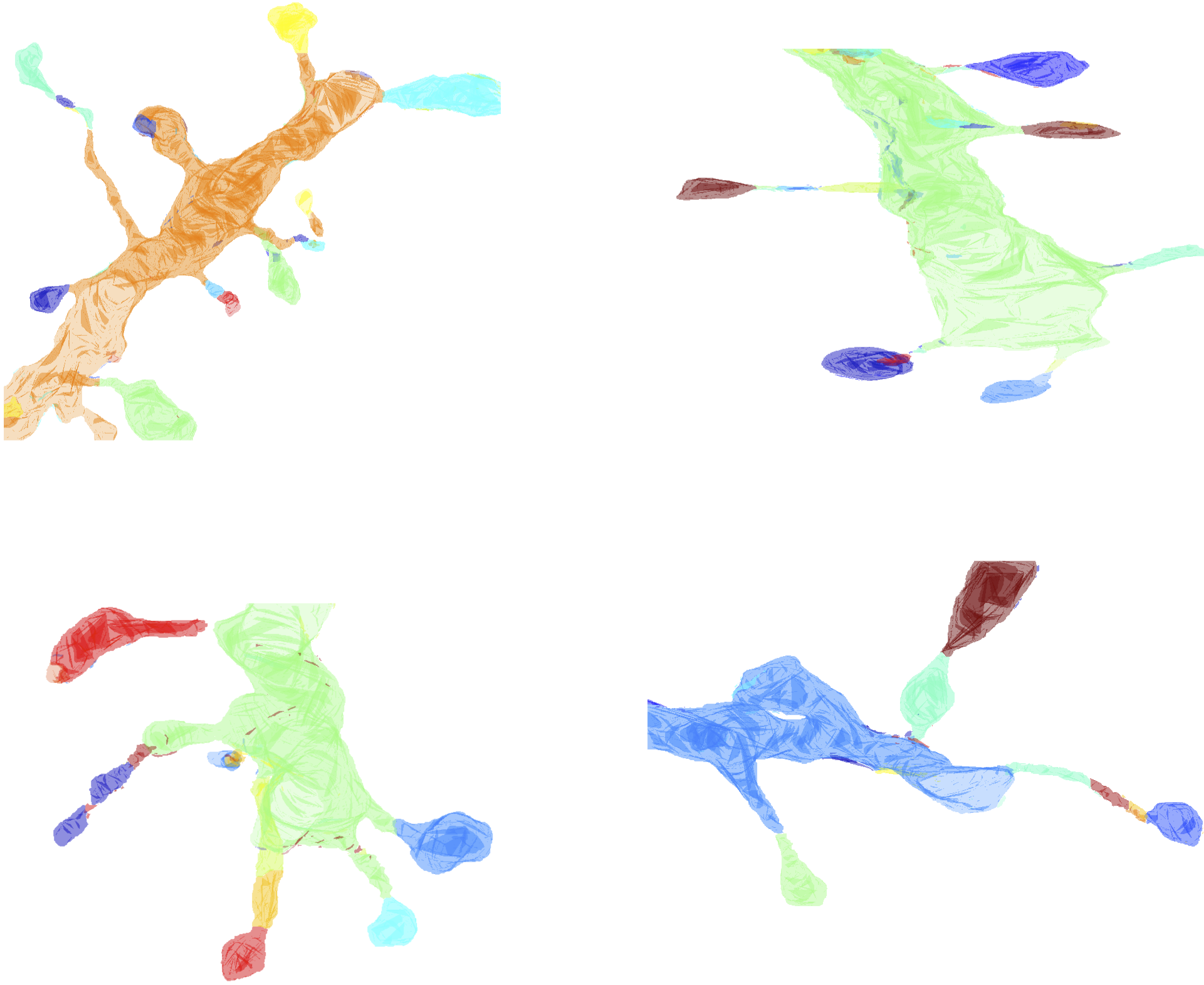}
		
    \end{minipage}%
    \begin{minipage}{0.35\textwidth}
        \centering
		\subfloat[slice 50]{\hspace{-0.2in}\includegraphics[width = 1.2in]{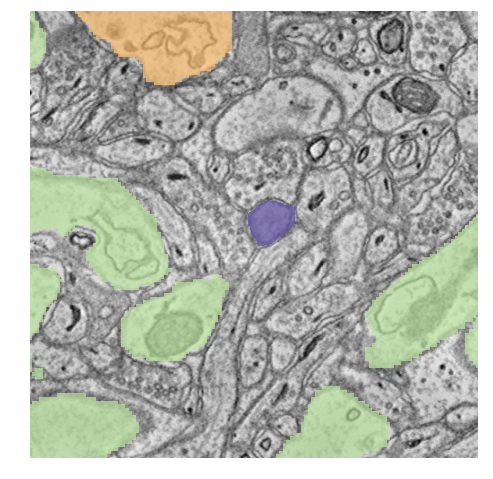}} 
		\subfloat[slice 60]{\hspace{.1in}\includegraphics[width = 1.2in]{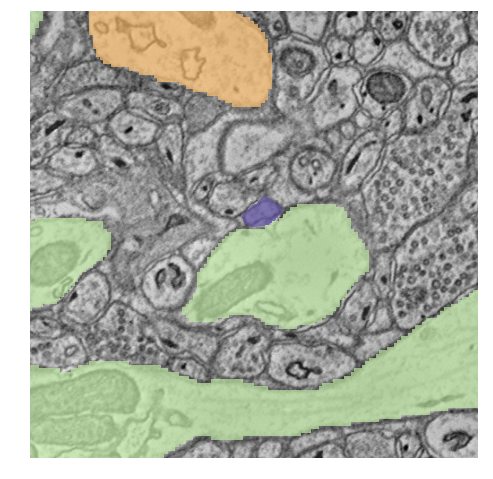}} \\ 
        \vspace{-.15in}
        
		\subfloat[slice 65]{\hspace{-0.2in}\includegraphics[width = 1.2in]{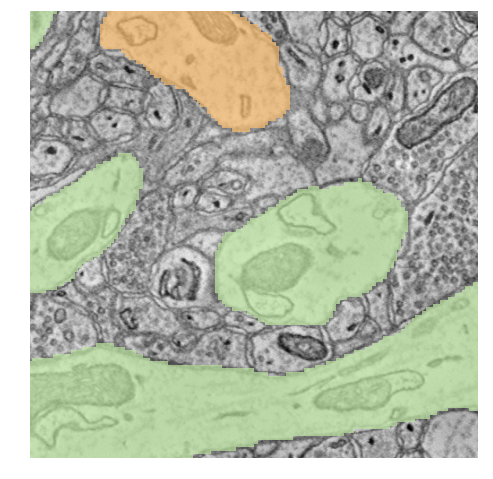}}
		\subfloat[slice 67]{\hspace{0.1in}\includegraphics[width = 1.2in]{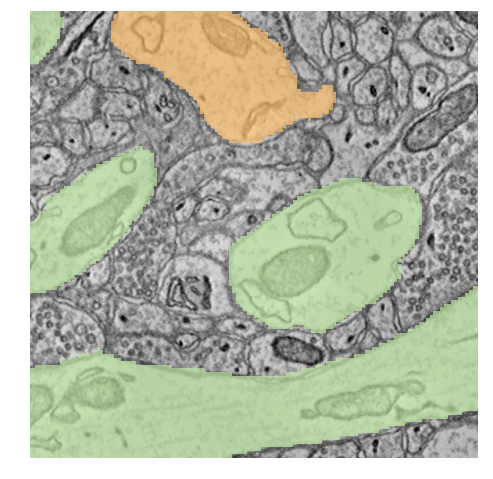}} 
        \vspace{-.15in}

		\subfloat[slice 69]{\hspace{-0.2in}\includegraphics[width = 1.2in]{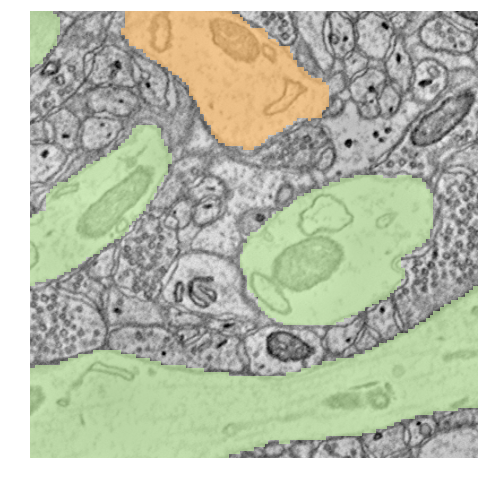}}       
        \subfloat[slice 73]{\hspace{0.1in}\includegraphics[width = 1.2in]{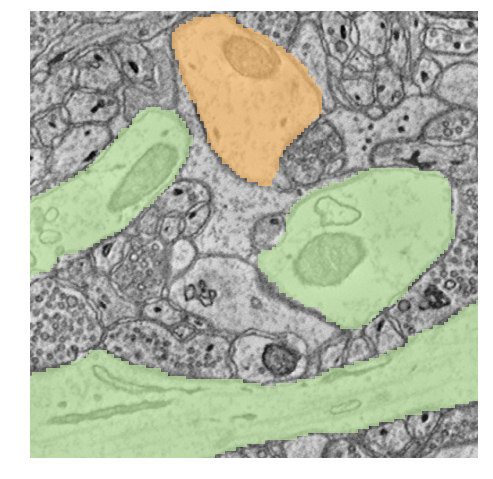}}\\ 
                
        \label{fig:spines}
    \end{minipage}
\caption{(Left) Examples of spine fragmentation.  Images illustrate typical split errors made in reconstructing spines by superimposing the automated segmentation labels on the ground truth for individual neurites. If reconstructed correctly, each object should be only a single color.  These illustrations actually understate the problem as they do not show merge errors for labels that extend beyond the ground truth mask. (Right) A typical spine merging problem --- spine is shown in blue but is incompletely linked; the true parent is in orange and other potential parent shafts are shown in green.}
\label{fig:overview}

\end{figure}

Traditionally, graphs are represented as $\mathcal{G} = (V,E,A)$, where vertices (V) represent each neuron, edges (E) represent connections or synapses between neurons, and both vertices and edges have attributes (A).  Recent work has shown that the network graph can be represented as a line graph in which synapses may be thought of as vertex terminals, and edges (i.e. neuron fragments) between them indicate pathways \cite{grayroncal2015}.  Although more complicated information is useful for downstream analysis (e.g., attributes like direction or weight), this basic connectivity question is perhaps the most fundamental of the unanswered connectomics questions.  Indeed,  when segmentation algorithms fail to connect these processes, the graph contains many disconnected nodes and an inaccurate picture of connectivity.  An illustration showing the challenges inherent in reconstructing spine-shaft linkages is shown in Figure~\ref{fig:overview}.

We believe that this is the first work to introduce an algorithm for solving the spine-shaft linking problem in serial section electron microscopy data.  However, several other groups have noted the difficulties associated with reconstructing dendritic spines and have developed semi-automated workflows to correct errors, including these spine fragments \cite{Mishchenko2010,Kaynig2013b,Helmstaedter2013}.  Of particular significance is research to assess and prioritize error proofreading based on connectivity impact \cite{Plaza2014}.  Other work has extensively studied dendritic spines (e.g., \cite{harris2007,Yuste2010}), providing a rich set of priors and information when reconstructing neuronal circuits.  Other   methods suggest related ideas, albeit from a different perspective or targeting a different setting \cite{Jackson2003,Turetken2013,Strandmark2015}.

In this work, we carefully explore prior EM segmentation results and develop an algorithm to reattach the fragmented spines, thus reconnecting many of the synapses that were previously graph isolates.  We leverage our understanding of local image grammars to develop a classifier that determines the best merge strategy for each spine.  We specifically focus on the spine-shaft problem, while acknowledging that related challenges such as synapse association, long range axonal projections, semantic typing, and merges across cuboid boundaries will need to be addressed when developing a comprehensive automated solution. We believe that this is the first algorithm to explicitly address the spine problem in the context of nano-scale connectomics, and so we hope that the datasets and methods presented here will serve as a testbed for future researchers. 

\section{Methods}

In vertebrate brains, anisotropic neuroimaging methods (e.g., electron microscopy) have the most difficulty resolving the finest processes \cite{Mishchenko2010}. When considering basic questions of connectivity, the spine necks are among the most important, yet most difficult to trace. 
Tracing large axons and dendrites may be possible at low resolution; however, resolving spine necks requires sub-10 nm resolution. 

\subsection{Grammars}

We observe that although the detailed wiring diagram of the brain is unknown, the high level structure of individual neurons obeys a predictable pattern.  This pattern is analogous to a tree having a trunk, branches, and leaves in a predictable arrangement.  Furthermore, although our datasets are large, the vocabulary of parts is constrained to only a few nouns, and local transitions between nouns can be considered independently of the surroundings, leading to a constrained, probabilistic context-free grammar. 
Note that automated reconstructions are frequently completed without considering higher order structure \cite{Nunez-Iglesias2013}. 

\begin{figure}[h!!]
\centering
\includegraphics[width=.6\textwidth]{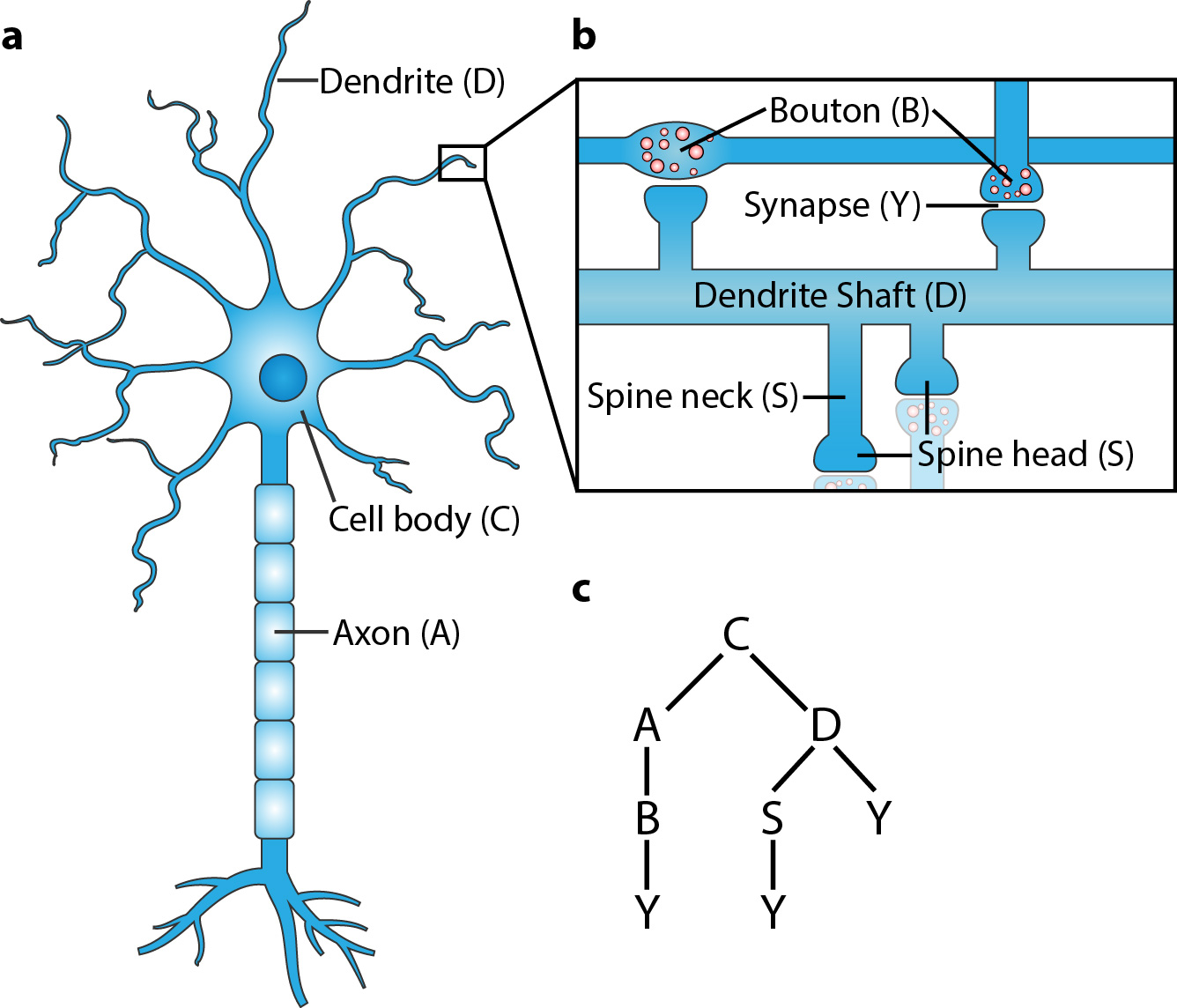}
\caption{Labeled parts of a neuron (A), with an inset showing our particular problem of interest (B).  A sample high-level parse tree capturing this information is shown for reference (C).}
\label{fig:geometry}
\end{figure}

At the highest level, neurons have the following parts: \textit{cell body (C)}, \textit{axon (A)}, \textit{dendrite}.  As described above, connections between neurons are especially important, and so we decompose the dendrite into two parts: \textit{dendritic spine (S)}, \textit{dendritic shaft (D)}, and define the additional symbols: \textit{axonal bouton (B)}, and \textit{synapse terminal (Y)} (Figure~\ref{fig:geometry}). Many other nouns are present in the neural tissue (e.g., glia, blood vessels, organelles), but these are not part of a basic graph  and so are not included in the production rules.  Self-transitions are possible due to branching or fragmentation in the reconstruction. Finally, because our initial experiments are of small volumes, comprising only pieces of individual neurons, all productions may also terminate at \textit{volume edges (E)}. \\

\noindent We express the basic grammar for a single neuron; these cellular units are built up to form a graph (with synapses as the connection points).  Termination points in an idealized grammar are always synapses; however in real datasets, cube boundaries are another possible termination symbol. Objects may be fragmented into multiple pieces, and so self-transitions are allowed but not shown here. Our grammar emphasizes topology rather than morphology, and is expressed in Backus-Naur Form:

\begin{verbatim}
C ::= A | D | Y | E
A ::= B | E 
D ::= S | Y | E 
B ::= Y | E
S ::= Y | E
\end{verbatim}

\vspace{0.25cm}

\noindent An example production (representing a normal flow of cellular information):

\begin{verbatim}
Y ::= B ::= A ::= C ::= D ::= S ::= Y  
\end{verbatim}

The same synaptic terminal (Y) will be shared by a second neuron.  In the vast majority of cases, we expect an axo-dendritic synapse motif, although other configurations are possible (e.g., axo-axonal, dendro-dendritic connections).  We do not observe these other patterns in our training data, but our grammar can be straightforwardly extended as needed. Next we examine two common production rules which are easy to exploit and the subject of this manuscript. The first represents the canonical connection between two cells; the second represents the connection between dendritic spines and shafts, which are the topic of this study.

\begin{verbatim}
A ::= B ::= Y =:: S | D
S ::= D 
\end{verbatim}

Although this paper targets the spine-dendrite production rule, we hope that this demonstration encourages the incorporation of other higher-level (biological) inference rules toward better circuit reconstructions.

\subsection{Measuring Error}

To date, success in connectomics has come from a combination of manual and automated processing \cite{White1986,Lu2009, Bock2011,Takemura2013}.  As imaging advances allow for the acquisition of ever larger data volumes, the reconstruction process becomes an expensive, enormously time-consuming bottleneck \cite{Marblestone2013}.  This challenge becomes even more daunting if one considers the potential variability in a single organism and that a full understanding of neuronal wiring diagrams likely requires the analysis of multiple organisms.  For these reasons, we continue to advocate for a fully automated intervention, with opportunities for semi-automated correction as required.  Recent research in graph theory suggests that even errorful graphs may still allow for the recovery of important neural motifs or primitives \cite{Priebe2013}.

We focus on two metrics leveraging the f-beta score \cite{Rijsbergen}. In this paper we fix $\beta = 1$, although we explored other values for estimating graph properties in internal experiments.  We first consider raw precision-recall scores of the spine-shaft association problem; this simply captures whether putative links are correct in an automated setting.  Second, we put these scores into a Top-K ranking setting, where we identify shafts that are likely partners for each spine.  This latter approach has applications for speeding up semi-automated proofreading workflows by allowing proofreaders to quickly choose from amongst a few choices rather than manually segmenting paths in an unconstrained environment.  This is an active area of research and promises to greatly impact circuit quality while improvements are made in fully automated algorithms  \cite{Plaza2014,Helmstaedter2013}.

\begin{equation}
f_\beta = \frac {(1 + \beta^2) \times \mathrm{true\ positive} }{(1 + \beta^2) \times \mathrm{true\ positive} + \beta^2 \times \mathrm{false\ negative} + \mathrm{false\ positive}}
\end{equation}

When reattaching spines, we also compute f1 graph error, which considers a brain graph to be like a communication network, where synapses are nodes and neuronal fragments represent the paths (i.e., connections) between them \cite{grayroncal2015}.  To compute this metric, we first construct a \textit{line graph} and then compute the harmonic mean of the precision and recall associated with these connections.  This measure is interpretible, because it converts graph error to a detection problem with false positive and false negative errors.

Although attributes like information direction or synaptic weight are useful for downstream analysis, the basic connectivity question we address here is perhaps the most fundamental.  The community has begun to focus on these questions as illustrated by the metrics posed by IARPA's MICrONS program\footnote{\url{https://www.iarpa.gov/index.php/research-programs/microns}} and the MICCAI CREMI challenge\footnote{\url{www.cremi.org}}.

\subsection{Pipeline}

\begin{figure}[h!!]
\centering
\includegraphics[width=\textwidth]{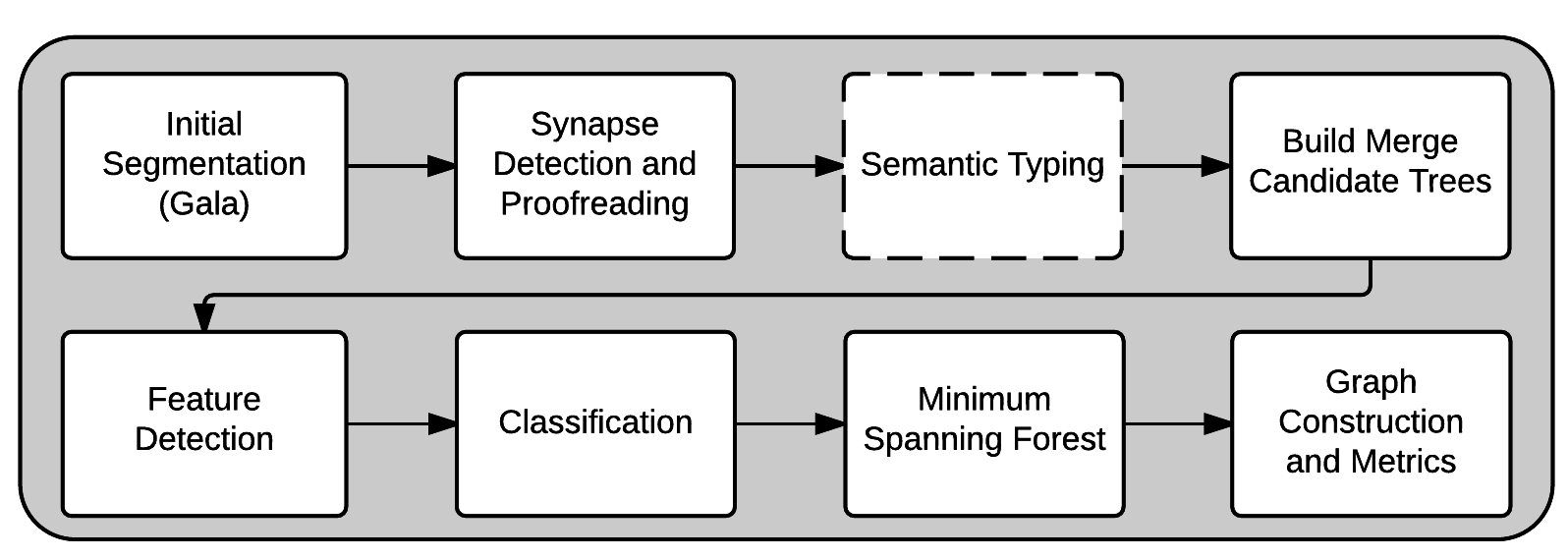}
\caption{A block diagram of our proposed approach for identifying fragmented spines.  We begin with a Gala segmentation and end with a graph.  Semantic typing is shown with a dashed line because this step is outside the scope of this paper.}
\label{fig:pipelineblock}
\end{figure}

We leverage the ideas developed above to guide our image processing and classification features to predict the best candidate shaft for each spine.  We first assess the characteristics of this problem and use them to develop a solution to improve the resulting network topology in an automated or semi-automated setting.  A block diagram of our approach is shown in Figure~\ref{fig:pipelineblock}.

\subsubsection{Data Preprocessing}

Our baseline method begins with known, semantically labeled shafts, spines, and synapses.  Estimating these labels is a problem that has been carefully studied, for neuron segmentation \cite{Funke2012,Nunez-Iglesias2013,Kaynig2013b,Parag2015}, synapse detection \cite{vesicle,Becker2013,Kreshuk2014},and semantically labeling objects \cite{vazquezreina2013,krasowksi2015}.  This work instead focuses on the linking problem that results after these algorithms have been run. 

More specifically, we begin with ground truth synapses and shafts, and use Gala \cite{vazquezreina2013} to segment the best spine candidate using agglomerative segmentation.  Gala is an often-used, high-performing \cite{grayroncal2015} technique that allows us to automatically generate a realistic estimate of the spine volume (reserving the spine truth information only for semantic labeling). Other segmentation methods may be used as inputs to \textit{Santiago} as the computed features are independent of any Gala specific metadata.  

\subsubsection{Spanning Trees}

Spines have a known distribution of distances between their head and the shaft \cite{Yuste2010}, which we exploit by looking for all shaft partners within a defined radius of each orphan spine.  As illustrated by the biological structure outlined above in Figure~\ref{fig:geometry}, each neuron has a tree structure, with each spine connected to exactly one parent.  There should be no orphan spines when neglecting boundary effects.  Therefore, we construct a spanning forest, consisting of a set of minimum spanning trees (one rooted at each spine).  To minimize computational complexity, we treat each spine orphan as an independent subproblem; the preserved spine-shaft link is the maximum probability edge in the graph.  Future versions of \textit{Santiago} could be extended to consider more complex interdependencies (e.g., periodic spine anchor locations, non-uniform spine distribution across dendrites).

We extract features by observing link distance, direction, and path cost. More specifically, we compute the following quantities:  minimum distance between spine and shaft; minimum distance between synapse and shaft; shaft size in window; minimum distance from end of spine to shaft; minimum distance from linearly propagated spine path; minimal path cost from spine to shaft (currently computed using membrane probabilities \cite{ciresan2012deep}); branching angle between spine and shaft.  These features are robust to a variety of settings and noise and provide an excellent estimation of the correct spine link.  For each feature  -- except spine-shaft branching angle -- we use both a raw score and relative ranking score to improve classifier robustness.  A version of \textit{Santiago} that uses only geometric label relationships could also be deployed to reduce data dependencies and speed processing.  

\subsubsection{Classification and assignment}

To determine the weight of each edge in a spanning tree (i.e., probability of a spine-shaft link), we use a random forest classifier composed of these features.  We follow a cross-validation approach, with all spines in the same dendritic parent group considered together (i.e., either in training or test) to minimize overfitting.  For each fold, we reserve one of these groups for testing and use the remaining groups to train the classifier.

Each link receives a probability score when applying the random forest classifier and we compute precision-recall on these links along with f-beta scores of 0.5, 1, and 2.  To compute a best overall graph-f1 score, we construct a spanning forest using a hard classification to predict the best candidate shaft for each spine.  When constructing a spanning tree, the associated edge weight is inversely related to these link probabilities (i.e. a high probability edge has a low link cost).

\section{Results}

We first develop data sets appropriate for characterizing and optimizing spine association.  We also carefully assess the impact of spines on overall connectivity, and demonstrate our algorithms on real data.

\subsection{Data}

Gold standard annotations for connectomes (especially in cortex) are still limited and challenging to leverage for automated analysis.  In this work we develop the first baseline dataset specifically designed to evaluate the spine problem, derived from a saturated, manual tracing in somatosensory cortex \cite{Kasthuri2015}.

\begin{figure}[htbp!!!]
\centering
\includegraphics[width=\textwidth]{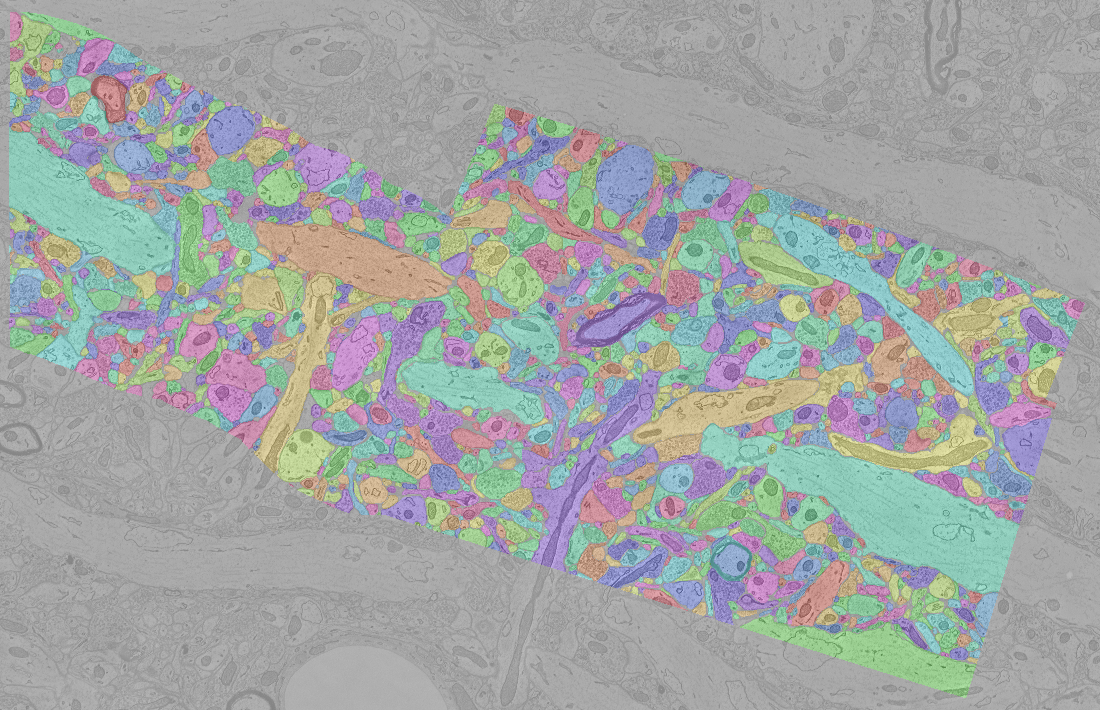}
\caption{A single slice of the primary segmentation (gold-standard) dataset used in this paper is shown above \cite{Kasthuri2015}.  Each color corresponds to a unique object (e.g, dendrite, glia, spine, axon).  Synapses are annotated in a different, spatially co-registered channel.} 

\label{fig:spines}
\end{figure}

\paragraph{3-cylinder dataset:} This dataset contains several thousand neurite fragments, 1700 synapses, and over 1000 spines. We curated this information to produce a dataset suitable for training and assessment.  To avoid conflating the spine assignment problem with earlier segmentation challenges, we work with data centered about a synapse with a biologically motivated cube size of about $2 \mu m$ in each direction (corresponding to $700 \times 700 \times 140$ voxel cuboids \cite{Yuste2010}.  Due to boundary conditions of the cylinder, some parts of these cutouts have no shafts and other shafts are cutoff; however, the resulting candidate merge trees provide a realistic, challenging scenario.  We partially mitigate these edge effects and spurious labels by admitting only objects explicitly labeled as spines by the original authors, and restrict shafts to large objects of at least a cubic micron.  Only those spines having a corresponding ground truth shaft parent in this restricted set are analyzed in this work.  These represent a significant fraction of the connections in the full data volume but eliminate other connections, which should be analyzed in future work.  This preprocessing procedure results in 531 spines and 38 target shafts for analysis.  The data used in this analysis can be explored online in a NeuroData \textit{ndviz} project.\footnote{ndviz: \url{http://viz.neurodata.io/project/kasthuri11_synapse_subcell/5/367/549/1100/}}

\paragraph{Images-to-graphs dataset:} Additionally a small region of this dataset containing reconstructions from part of the `AC3 region' were used in a recent analysis to assess graph-f1 error \cite{grayroncal2015}.  We use this data in our simulations to assess the impact of fragmented synaptic connections to partially justify the importance of this work.  Because this volume is small (spanning $1024 \times 1024 \times 100$ pixels, some edge effects in labeling are also present in these data, as processes transiting the edges of the volume can lead to association information unavailable to an automated algorithm.

\subsection{Infrastructure}

To perform these experiments, we leveraged the NeuroData\footnote{NeuroData: \url{neurodata.io}} infrastructure.  To extract various cubes of data efficiently and repeatably, we used ndio \cite{matelsky2016}\footnote{ndio: \url{github.com/neurodata/ndio}}, which implements a data access API for image and annotation data.  

\subsection{Simulation Results}

As others \cite{Kaynig2013b,Plaza2014,grayroncal2015} have noted, spines are a major issue in graph connectivity, but to date, the  impact has not been quantified in the context of electron microscopy graph estimation.  Here we explore a quantitative assessment through simulation.  

\subsubsection{3-Cylinder Dataset}

\begin{figure}[htbp!!!]
\centering
\includegraphics[width=\textwidth]{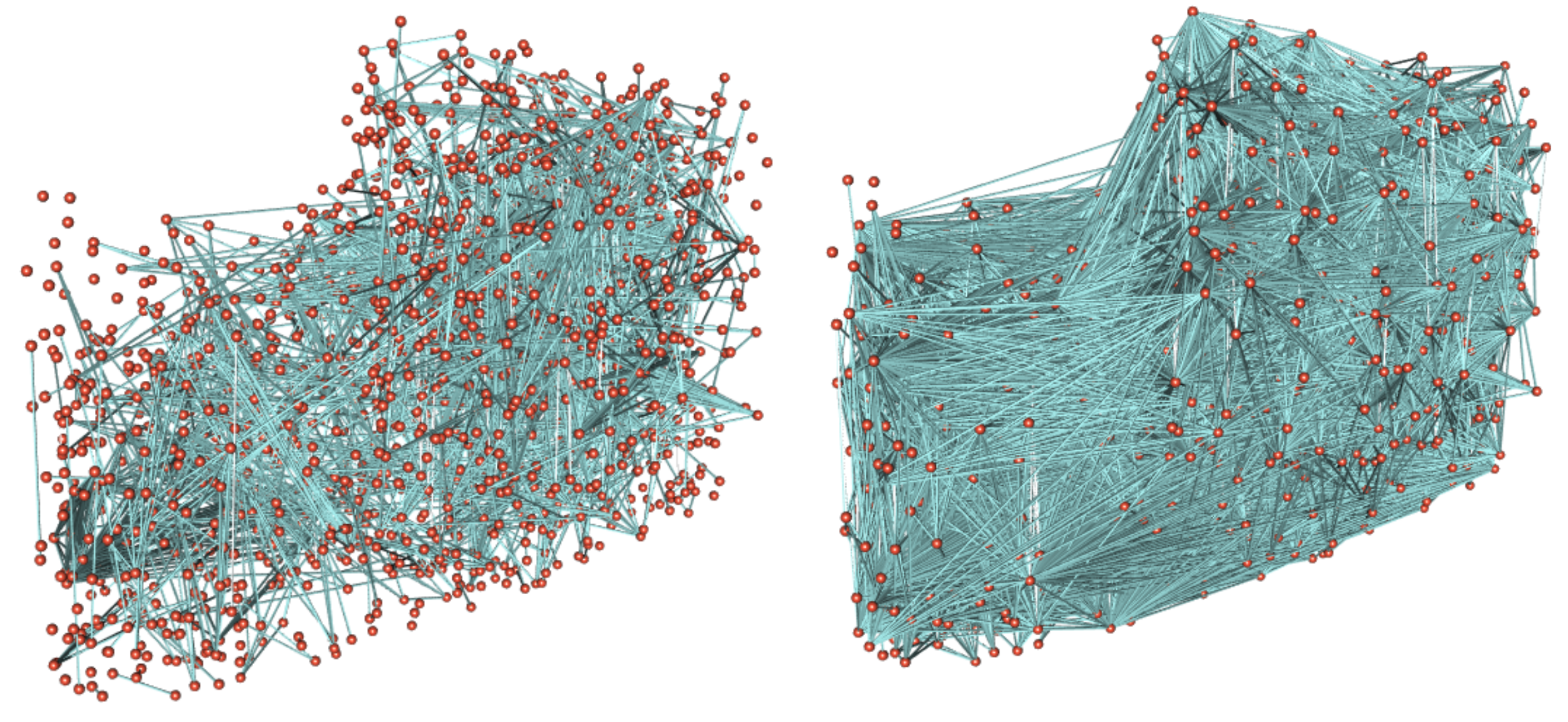}
\caption{Graph with disconnected spines (left) and gold standard graph (right).  These illustrations emphasize the large impact of spines on the overall graph connectivity.}
\label{fig:impact}
\end{figure}

We begin with the true connectivity matrix where all spines are correctly associated with their parents in the 3-cylinder dataset.  We investigate the impact of spine-shaft linking on overall connectivity (i.e., the line graph), by detaching spines.  This process creates a separate segmentation label for the spine, leading to a synapse that is effectively disconnected on the dendritic side.  Because high degree nodes impact the graph disproportionately, we repeat our simulations at different levels of spine fragmentation, quantifying error, average degree and their variances.  In the true line graph, the total number of edges is  31,980 (average degree: 18.8).  In the graph with all spines disconnected, the total number of edges is 2,718 (average degree: 1.6); a vizualization highlighting these differences can be seen in Figure~\ref{fig:impact}.

\begin{figure}[htbp!!!]
  \centering
  \includegraphics[width=\textwidth]{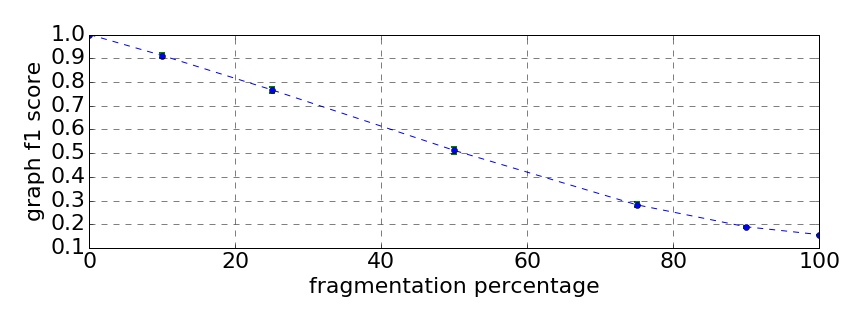}

  \caption{Graph error as a function of spine fragmentation (0-100\%) showing the f1 graph error \cite{grayroncal2015}.  This firmly establishes the importance of spines on connectivity, especially at a local scale.  1000 iterations were performed with different spines removed each time. }
  \label{fig:3cylscore}
\end{figure}

For the 3-cylinder dataset, we examine this spine fragmentation and conclude that spine association is necessary but not sufficient to ensure an accurate connectome -- most of the connections are carried on spines and are lost when these spines are fragmented. We show this quantitatively in Figure~\ref{fig:3cylscore}.  

We further explored this idea by running Gala on a region surrounding each spine and found that nearly all (86\%) of the spine-shaft linkages were missed.  A successful match was scored whenever the most common segmentation label for the spine truth and shaft truth was identical; this assessment disregards overmerging failures.  Our cutout region was a $700 \times 700 \times 140$ voxel window, corresponding to a cutout of $4 \times 4 \times 4 \,\mu m$, centered about the synapse of interest.  In our anisotropic data, this was sufficient to capture nearly all (97\%) of the shaft partners for the spines of interest.   

\subsubsection{Images-to-Graphs Dataset}

To further understand the spine problem, we conduct an additional simulation using the  images-to-graphs dataset.  We identify all synapse orphans as putative spines in a Gala segmentation (threshold: 0.5, tuned to reduce the possibility of overmerging).  Because no semantic labels are available, we select all orphans incident to a synapse.  We then synthetically merge those objects to their parent ``shaft," which we select as the largest object in the dataset with the same truth label as computed through overlap with corresponding ground truth labels. 

The resulting score improves the baseline f1-graph score of 0.31 to 0.64.  This again emphasizes the importance of spines, as we can  double the graph-f1 score by identifying and linking these orphans without altering the other segmentations.  This result suggests that existing algorithms are accurate at reconstructing large processes and by focusing on these small, disproportionately important objects that violate a known biological constraint (i.e. connectedness), we can address many of the deficiencies of conventional algorithms with a hierarchical approach. 

\subsection{Computer Vision Results}
\FloatBarrier
\begin{table}[htbp!!]
\centering
\caption{Table showing computer vision results on each dataset. Baseline prior to this algorithm is zero matches, as we operate only on spines that are missed by Gala.}
\label{table:cv1}
\begin{tabular}{|l|c|}
\hline
\textbf{3-Cylinder Kasthuri Dataset \cite{Kasthuri2015}} & \textbf{Scores} \\ \hline
\textbf{Top-1 (match)} & 203 / 455 = 0.45 \\
\textbf{Top-2} & 299 / 455 = 0.66 \\
\textbf{Top-3}  & 352 / 455 = 0.77\\
\textbf{Top-5}  & 405 / 455 = 0.89 \\
\textbf{Top-10} & 436 / 455 = 0.96 \\
\textbf{Maximum f1} & 0.47 \\ 
\textbf{Median rank (when available)} & 2 \\
\textbf{Mean rank (when available)} & 2.41 \\ \hline
\textbf{Spines with truth in window} & 440 / 445 = 0.97\\ 
\textbf{Average shafts / spanning tree} & 9.5 \\ \hline
\end{tabular}
\end{table}

We apply our spines-shafts pipeline to identify candidate shafts for each orphan spine. In this work we leverage previously computed membranes \cite{grayroncal2015} and compute Gala segmentations on a small region surrounding each orphan synapse (as discussed above).  Running Gala takes about 2 hours and 20GB of RAM on a single core for this sized volume. NeuroProof,\footnote{NeuroProof: \url{https://github.com/janelia-flyem/NeuroProof}} a successor to Gala,  offers faster computation and additional options, but was not specifically evaluated for this work.  Computing features and classification took approximately 15 minutes per spine; the bulk of this time was spent computing path finding features.  

In all of the results reported below, we only show orphan spines (excluding the 76 Gala spines that successfully matched).  Therefore, prior work is that all of these spines are disconnected from their parents (0/455).  We note that a few (15/455) spines do not have shafts present due to the window size chosen; these are not excluded and are treated as errors when reporting algorithm performance.

In Table~\ref{table:cv1} we provide a detailed reporting of results showing  conventional f1-detection metrics for both automated processing of edges (e.g., maximum f1) and Top-K performance, for use in semi-automated proofreading approaches.  We also report our mean and median ranks when the true parent shaft is present and the number of available shafts available on average for each linking scenario.  

In Table~\ref{table:cv2}, we report the results of our algorithm on graph-f1 error.  We show two columns of values: the \textit{Santiago} subgraph contains only the connections and their immediate partners used in our test dataset (i.e., the axon and dendrite fragment and corresponding parent neuron information); the 3-cylinder graph contains all connections and demonstrates that fixing errors in the local subgraph translates to overall graph quality improvement.  The spanning forest results shows our best overall automated performance, while the Top-K results show the simulated impact of humans successfully (perfectly) proofreading the Top-K results.  In a real-world proofreading setting, a user could be presented with options and would either identify the true partner or return no match.  As K increases, the operator workload (and potential performance) will correspondingly increase.

\begin{table}[htbp]
\centering
\caption{f1 graph scores on \textit{Santiago} subgraph and full 3-cylinder graphs.  The table below shows a baseline for performance prior to \textit{Santiago} being run and afterwards.  The post-run numbers include an assessment using fully-automated and semi-automated approaches.}
\label{my-label}
\begin{tabular}{|l|c||l|c|}
\hline
\textbf{\textit{Santiago} subgraph} & \textbf{f1 graph score} & \textbf{3-cylinder graph} & \textbf{f1 graph score} \\ \hline
no spines & 0.035 & no spines & 0.205 \\
gala spines only & 0.089 & gala spines only & 0.246 \\
spanning forest (auto) & 0.404 & spanning forest (auto) & 0.398 \\
top-1 (proofread) & 0.518 & top-1 (proofread) & 0.462 \\
top-2 (proofread) & 0.711 & top-2 (proofread) & 0.573 \\
top-3 (proofread) & 0.816 & top-3 (proofread) & 0.638 \\
top-5 (proofread) & 0.910 & top-5 (proofread) & 0.701 \\
all (proofread) & 0.983 & all (proofread) & 0.752 \\ \hline
\end{tabular}
\label{table:cv2}
\end{table}
\vspace{0.5cm}
\section{Discussion}

In this paper, we reframe the connectomics problem to focus explicitly on connectivity, and measure progress using a convenient, descriptive metric.  We illustrate a semantic, biologically inspired solution to partially remedy one of the major problems of neuron reconstruction (i.e., linking spines to dendritic shafts).  By inferring paths that may not be clear at a voxel-level, we are able to recover connections that would otherwise have been lost.  The field of connectomics is still in its infancy; this work provides an early example of the untapped potential for combining well-studied biological phenomena to computer vision approaches.  Although we demonstrate this idea in the context of electron microscopy and a particular segmentation algorithm, the underlying principles are very general and potentially could be leveraged in other settings including light microscopy where spine necks are unable to be resolved due to resolution constraints.   The tree structure and biological priors provide a scaffold that constrains the reconstruction puzzle and may greatly facilitate both estimation and error-checking as models improve.

We emphasize the importance of connection, rather than segmentation, and propose a new solution that allows for many spines to be recovered that are missed using conventional approaches.  Our algorithm is agnostic to the segmentation ``preprocessing'' method and will likely improve as coarse segmentations improve. Future work will apply these techniques to different data sets and problem settings and also explore fully-automated approaches (e.g. including semantic typing) and integration into a complete pipeline. Our preprocessing segmentation algorithm, Gala, missed most of the spines, requiring \textit{Santiago} to do extensive reassembly.  As segmentation algorithms improve or incorporate these biological priors directly, the post-processing  required may be substantially reduced, improving connectome fidelity.

Finally, we produced a database of spines and shafts which enable future algorithm development and testing.  Our methods and approach are scalable and fit into a broader effort that seeks to transform images into graphs.  Our code and data are publicly available in accordance with reproducible science.\footnote{Paper webpage: \websitesantiago}

\section*{Acknowledgments}

The authors wish to thank Dean Kleissas and Nathan Drenkow for helpful discussions on the spine-shaft problem; Alex Baden, Kunal Lillaney, Randal Burns and Joshua Vogelstein for NeuroData support; and Greg Kiar for rendering the graphs appearing in this manuscript.

\FloatBarrier
\newpage

\bibliographystyle{IEEEtran}
\bibliography{santiago}	

\end{document}